\documentclass[a4paper,twoside]{article}

\usepackage{epsfig}
\usepackage{subcaption}
\usepackage{calc}
\usepackage{amssymb}
\usepackage{amstext}
\usepackage{amsmath}
\usepackage{amsthm}
\usepackage{multicol}
\usepackage{pslatex}
\usepackage{apalike}
\usepackage{algorithm2e}
\usepackage{booktabs}
\usepackage{graphicx}
\usepackage{array}

\usepackage[bottom]{footmisc}
\usepackage{SCITEPRESS}     

\usepackage{makecell}
\usepackage{xcolor}
\usepackage{url}
\usepackage{siunitx}
\usepackage[utf8]{inputenc}

\usepackage{hyperref}
\definecolor{darkblue}{rgb}{0, 0, 0.5}
\hypersetup{colorlinks=true, citecolor=black, linkcolor=black, urlcolor=darkblue}

\usepackage{fancyhdr}

\fancypagestyle{firstpage}{
  \fancyhf{} 
  \fancyfoot[C]{\thepage} 
  \fancyfoot[L]{\scriptsize Accepted at ICAART 2025. Official version: \href{https://doi.org/10.5220/0013374100003890}{doi.org/10.5220/0013374100003890}} 
}

\begin{document}

\title{Large Language Models for Summarizing Czech Historical Documents and Beyond}
\thispagestyle{firstpage} 

\author{\authorname{
V\'{a}clav Tran\sup{1}\orcidAuthor{0009-0003-0250-2821},
Jakub \v{S}m\'id\sup{1, 2}\orcidAuthor{0000-0002-4492-5481},
Ji\v{r}\'{i} Mart\'{i}nek\sup{1, 2}\orcidAuthor{0000-0003-2981-1723},
Ladislav Lenc\sup{1, 2}\orcidAuthor{0000-0002-1066-7269}
and Pavel Kr\'{a}l\sup{1, 2}\orcidAuthor{0000-0002-3096-675X}}
\affiliation{\sup{1}Department of Computer Science and Engineering, University of West Bohemia in Pilsen, Univerzitn\'{i}, Pilsen, Czech Republic}
\affiliation{\sup{2}NTIS - New Technologies for the Information Society, University of West Bohemia in Pilsen, Univerzitn\'{i}, Pilsen, Czech Republic}
\email{nuva@students.zcu.cz, \{jaksmid, llenc, jimar, pkral\}@kiv.zcu.cz}
}


\abstract{
Text summarization is the task of shortening a larger body of text into a concise version while retaining its essential meaning and key information. While summarization has been significantly explored in English and other high-resource languages, Czech text summarization, particularly for historical documents, remains underexplored due to linguistic complexities and a scarcity of annotated datasets. Large language models such as Mistral and mT5 have demonstrated excellent results on many natural language processing tasks and languages. 
Therefore, we employ these models for Czech summarization, resulting in two key contributions: (1) achieving new state-of-the-art results on the modern Czech summarization dataset SumeCzech using these advanced models, and (2) introducing a novel dataset called Posel od \v{C}erchova for summarization of historical Czech documents with baseline results. Together, these contributions provide a great potential for advancing Czech text summarization and open new avenues for research in Czech historical text processing.
}

\onecolumn \maketitle \normalsize \setcounter{footnote}{0} \vfill

\section{\texorpdfstring{\uppercase{Introduction}}{Introduction}}
\label{sec:introduction}
The rapid evolution of Natural Language Processing (NLP) techniques has elevated the performance of text summarization systems. While most advances focus on high-resource languages like English, the Czech language, particularly historical variations, remains underrepresented. Historical Czech documents pose unique challenges due to linguistic shifts, outdated vocabulary, and inconsistent syntax. These nuances create a significant gap in the development of automated summarization systems capable of handling this domain effectively.

Therefore, this paper addresses two interlinked challenges. First, it seeks to establish new state-of-the-art benchmarks on SumeCzech, the most comprehensive dataset for modern Czech text summarization using modern Large Language Models (LLMs), namely Mistral~\cite{jiang2023mistral} and mT5~\cite{xue-etal-2021-mt5}. Second, recognizing the lack of resources tailored for historical Czech, we introduce a newly created dataset derived from the historical journal Posel od \v{C}erchova. The dataset is specifically designed to facilitate summarization tasks in historical contexts, enabling future researchers to address the linguistic complexities inherent in this domain. This corpus is freely available for research purposes\footnote{\href{https://corpora.kiv.zcu.cz/posel\_od\_cerchova/}{corpora.kiv.zcu.cz/posel\_od\_cerchova/}}. 

By combining model advancements and dataset innovation, this research aims to drive progress in the Czech summarization field and open venues for applications in cultural preservation, historical research, and digital humanities.

\section{\texorpdfstring{\uppercase{Related Work}}{Related Work}}
Text summarization methods can be categorized into abstractive and extractive ones. Extractive summarization selects the most representative sentences from the source document, while abstractive summarization generates summaries composed of newly created sentences.

Early summarization methods were extractive ones and relied on statistical and graph-based methods like TF-IDF (Term Frequency-Inverse Document Frequency)~\cite{ChristianTFIDFSumm2016}, which scores sentence importance based on term frequency relative to rarity across a corpus. 
Similarly, TextRank~\cite{mihalcea-tarau-2004-textrank} represents sentences as nodes in a graph and ranks them using the PageRank algorithm~\cite{Page1999ThePC}. 

Neural networks advanced both extractive and also abstractive summarization by modeling sequences with Recurrent Neural Networks (RNNs)~\cite{elman1990finding}. One extractive approach involves sequence-to-sequence architectures where LSTM models capture the contextual importance of each sentence within a document~\cite{nallapati2017summarunner}. 
Hierarchical attention networks combine sentence-level and word-level attention to better capture document structure and relevance for summarization~\cite{yang2016hierarchical}. This approach has proven effective in summarizing longer and more complex documents. 
Hybrid approaches combining BERT embeddings~\cite{devlin2019bert} with K-Means clustering~\cite{kmeans} to identify key sentences~\cite{miller2019leveraging} have shown excellent performance for abstractive summarization. 

Advances in sequence-to-sequence Transformer-based models~\cite{vaswani2017attention} have revolutionized abstractive summarization. Recent models like T5~\cite{2020t5} adopt a text-to-text framework and excel in various tasks, including summarization, due to pre-training on the C4 dataset. PEGASUS~\cite{zhang2019pegasus} introduces gap sentences generation for masking key sentences during pre-training, achieving strong performance on 12 datasets. Similarly, BART~\cite{lewis2019bart} uses denoising objectives for robust text summary generation. Multilingual models such as mT5~\cite{xue-etal-2021-mt5} and mBART~\cite{liu-etal-2020-multilingual-denoising} extend these capabilities to multiple languages, including Czech, through datasets like mC4~\cite{xue-etal-2021-mc4} and multilingual Common Crawl\footnote{http://commoncrawl.org/}.

However, these models often underperform on non-English corpora without fine-tuning.

\section{\texorpdfstring{\uppercase{Datasets}}{Datasets}}
The following section provides a brief review of the primary existing summarization datasets.
Moreover, the created {Posel od \v{C}erchova} corpus will also be detailed at the end of this section.

\subsection{English Datasets}
{\bf CNN/Daily Mail}~\cite{hermann2015cnndm} dataset consists of over 300,000 English news articles, each paired with highlights written by the article authors. It has been widely used in summarization and question-answering tasks, evolving through several versions tailored for specific NLP tasks.

{\bf XSum}~\cite{xsumDataset} contains 226,000 single-sentence summaries paired with BBC articles covering diverse domains such as news, sports, and science. Its focus on single-sentence summarization makes it less biased toward extractive methods.

{\bf Arxiv Dataset}~\cite{cohan-etal-2018-discourse} includes 215,000 pairs of scientific papers and their abstracts sourced from arXiv. It has been cleaned and formatted to ensure standardization, with sections like figures and tables removed.

{\bf BOOKSUM}~\cite{kryscinski-etal-2022-booksum} is a dataset tailored for summarizing long texts like novels, plays, and stories, with summaries provided at paragraph, chapter, and book levels. Texts and summaries were sourced from Project Gutenberg and other web archives, supporting both extractive and abstractive summarization.

\subsection{Multilingual Datasets}
{\bf XLSum}~\cite{xlsumDataset} provides over one million article-summary pairs across 44 languages, ranging from low-resource languages like Bengali and Swahili to high-resource languages such as English and Russian. Extracted from various BBC sites, this dataset is a valuable resource for multilingual summarization research.

{\bf MLSUM}~\cite{mlsumDataset} consists of 1.5 million article-summary pairs in five languages: German, Russian, French, Spanish, and Turkish. The dataset was created by archiving news articles from well-known newspapers, including Le Monde and El Pais, with a focus on ensuring broad topic coverage.

The above-mentioned datasets are for English summarization, and some are multilingual; however, Czech resources remain very limited. 

\subsection{SumeCzech}
SumeCzech large-scale dataset~\cite{straka2018sumeczech} is a notable exception to the scarcity of Czech-specific resources. This dataset was created at the Institute of Formal and Applied Linguistics at Charles University and is tailored for summarization tasks in the Czech language. It comprises one million Czech news articles. These articles are sourced from five major Czech news sites: České Noviny, Deník, iDNES, Lidovky, and Novinky.cz. Each document is structured in JSONLines format, with fields for the URL, headline, abstract, text, subdomain, section, and publication date. The preprocessing includes language recognition, duplicate removal, and filtering out entries with empty or excessively short headlines, abstracts, or texts.

This dataset supports multiple summarization tasks, such as headline generation and multi-sentence abstract generation. The training, development, and testing splits are in roughly 86.5/4.5/4.5 ratio. 
The average word count is 409 for full texts and 38 for abstracts.

 Nevertheless, this dataset caters exclusively to modern Czech and fails to address the needs of historical text processing.

\subsection{Posel od \v{C}erchova}
 To construct the dataset, we used data from the historical journal \textit{Posel od \v{C}erchova (POC)}, which is available on the archival portal Porta fontium\footnote{\url{https://www.portafontium.eu}}.
 
The construction of the dataset involved addressing the challenge of creating summaries for the provided texts, which were composed in historical Czech and, in some rare cases, even German. The texts also covered a variety of different topics, from local news surrounding Domažlice (a historic town in the Czech Republic), opinion pieces, and various local advertisements to internal and worldwide politics and feuilletons. Furthermore, it was important to construct a dataset of sufficient size to ensure the accuracy and reliability of the evaluation. These aspects added complexity to the summarization task.

To overcome the mentioned issues, we employed state-of-the-art (SOTA) LLMs GPT-4~\cite{openai2024gpt4} and Claude 3 Opus~\cite{anthropic-2024} (Opus) (specifically the \texttt{claude-3-opus-20240229} version) for initial text summary creation. These models were selected based on their SOTA performance in many NLP tasks and excellent performance in some preliminary summarization experiments.

While generating the summaries, it was essential to ensure conciseness. Since most of the implemented methods were fine-tuned on the SumeCzech dataset, we aimed to maintain consistency by creating summaries in a journalistic style, reflecting the dataset's characteristics. To achieve this, the prompts for generating the summaries included explicit instructions, as shown below:

\begin{itemize}
\item Vytvoř shrnutí následujícího textu ve stylu novináře. Počet vět $<= 5$; (EN: Create a summary of the following text in the style of a journalist. Number of sentences $<= 5$)
\end{itemize}

During the summarization task, we observed that while both models produced summaries of very good quality, Opus tended to create more succinct and stylistically appropriate ones, closely aligning with the news reporter format. However, there were instances where summaries generated by Opus exhibited an excessive focus on a single topic. 

On the other hand, GPT-4 aimed to incorporate a greater level of detail within the five-sentence constraint but occasionally deviated from the specified stylistic prompt. 

If the model-generated summary exhibited significant stylistic deviations or excessive focus on a single topic, we either modified or regenerated it until a correct version was achieved.

Two-level summaries were created; the first one was on the page level, and the second one summarizes a whole article that is usually composed of several pages.
We thus summarized 432 pages, effectively resulting in the creation of 100 issue summaries.
The subset containing page summaries is hereafter referred to as {\it POC-P}, while the issue summaries are referred to as {\it POC-I}.
Note that all created summaries were checked and corrected manually by two native Czech speakers.

The dataset is in the {\bf .json} format and contains the following information:

\begin{itemize}
    \item \textbf{text:} Text extracted from the given page, a digital rendition of the original printed content;
    \item \textbf{summary:} Summary of the page, which is no more than 5 sentences long;
    \item \textbf{year:} Publication year of the journal;
    \item \textbf{journal:} Specification of the source journal: the day, month, and the number of the issue is contained within this identifier;
    \item \textbf{page\_src:} Name of the source image file converted into the text;
    \item \textbf{page\_num:} Page number.
\end{itemize}

This dataset is designed to support summarization tasks within Czech historical contexts, providing researchers with the tools to tackle the linguistic challenges unique to this domain. The corpus is freely accessible for research purposes\footnote{https://corpora.kiv.zcu.cz/posel\_od\_cerchova/}.

\section{\texorpdfstring{\uppercase{Models}}{Models}}
The experiments employ two advanced Transformer-based models, Multilingual Text-to-Text Transfer Transformer (mT5)~\cite{xue-etal-2021-mt5} and Mistral 7B~\cite{jiang2023mistral}.

\subsection{Multilingual Text-to-Text Transfer Transformer}
The Multilingual Text-to-Text Transfer Transformer (mT5) is a variant of the T5 model designed for multilingual tasks. This model is trained on the multilingual mC4 dataset~\cite{xue-etal-2021-mc4}, which includes Czech, and effectively handles a wide range of languages. The model is based on Transformer encoder-decoder architecture and uses a SentencePiece tokenizer~\cite{kudo-richardson-2018-sentencepiece} to process complex language structures, including Czech morphology. Pre-trained using a span corruption objective~\cite{raffel2020t5}, mT5 predicts masked spans of text, enabling it to learn semantic and contextual relationships.

The mT5 model is available in various sizes, from small with 300 million parameters to XXL with 13 billion parameters, and is therefore adapted to different computational needs. The base variant of the mT5, which contains 580 million parameters, is used for further experiments.

\subsection{Mistral Language Model}
The Mistral Language Model (Mistral LM) is a highly efficient large language model known for its robust performance across diverse natural language processing tasks. It is designed to combine high accuracy with computational efficiency, achieving state-of-the-art results in reasoning, text generation, summarization, and other NLP applications. Mistral 7B, with its 7 billion parameters, strikes a balance between computational efficiency and task performance, surpassing larger models like 13B or 34B in several benchmarks.

This model utilizes advanced attention mechanisms like Grouped-Query Attention (GQA)~\cite{ainslie2023gqa} and Sliding Window Attention (SWA)~\cite{beltagy2020longformer}. GQA enhances processing speed by grouping attention heads to focus on the same input data, while SWA reduces computational costs by limiting token attention to nearby tokens. The model supports techniques such as quantization~\cite{gholami2021survey} and Low-Rank Adaptation (LoRA)~\cite{hu2021lora} for efficient fine-tuning on limited hardware, enabling it to handle longer inputs effectively.

\section{\texorpdfstring{\uppercase{Experiments}}{Experiments}}

\subsection{Evaluation Metrics}
The following evaluation metrics are used.

ROUGE (Recall-Oriented Understudy for Gisting Evaluation)~\cite{lin-2004-rouge} is a set of metrics used to evaluate the quality of summaries by comparing n-gram overlaps between a system-generated summary and reference texts. Key ROUGE metrics include ROUGE-N (for n-gram overlap) and ROUGE-L (for the longest common subsequence).

ROUGERAW~\cite{straka2018rougeraw} is a variant of ROUGE that evaluates raw token-level overlaps between predicted and reference texts without any preprocessing like stemming or lemmatization. It measures exact matches of tokens, making it suitable for tasks where precise token alignment is important.

\subsection{Set-up}
 We used AdamW optimizer~\cite{loshchilov2017decoupled} with a learning rate set to 0.001 as suggested by authors of mT5~\cite{xue-etal-2021-mt5} for the training of this model. 
For Mistral 7B, we utilized QLoRA~\cite{qlora}, a method that integrates a 4-bit quantized model with a small, newly introduced set of learnable parameters. During fine-tuning, only these additional parameters are updated while the original model remains frozen, thereby substantially reducing memory requirements. We employ the models from the HuggingFace Transformers library~\cite{wolf-etal-2020-transformers}.
 For training both models, we used a single NVIDIA A40 GPU with 45 GB VRAM.

\subsection{Model Variants}
We use three variants of the models in our experiments: 
\begin{itemize} 
    \item M7B-SC: The Mistral 7B model fine-tuned on the SumeCzech dataset; 
    \item M7B-POC: The Mistral 7B model further fine-tuned on the POC dataset; 
    \item mT5-SC: The mT5 model fine-tuned on the SumeCzech dataset. 
\end{itemize}
 
 \begin{table*}[ht!]
    \centering
    \caption{Results of various methods on SumeCzech dataset with precision (P), recall (R), and F1-score (F).}
    \label{tab:eval:sc_scores_comparison}
    \begin{tabular}{
        l
        r
        r
        r
        r
        r
        r
        r
        r
        r
    }
        \toprule
        Method & \multicolumn{3}{c}{ROUGE\textsubscript{raw}-1} & \multicolumn{3}{c}{ROUGE\textsubscript{raw}-2} & \multicolumn{3}{c}{ROUGE\textsubscript{raw}-L} \\
        \cmidrule(lr){2-4} \cmidrule(lr){5-7} \cmidrule(lr){8-10}
        & {P} & {R} & {F} & {P} & {R} & {F} & {P} & {R} & {F} \\
        \midrule
        M7B-SC & \textbf{24.4} & \textbf{19.7} & \textbf{21.2} & \textbf{6.5} & \textbf{5.3} & \textbf{5.7} & \textbf{17.8} & \textbf{14.5} & \textbf{15.5} \\
        mT5-SC & 22.0 & 17.9 & 19.2 & 5.3 & 4.3 & 4.6 & 16.1 & 13.2 & 14.1 \\
        \midrule
        HT2A-S~\cite{Krotil2022TextSummarization} & 22.9 & 16.0 & 18.2 & 5.7 & 4.0 & 4.6 & 16.9 & 11.9 & 13.5 \\
        First~\cite{straka2018sumeczech} & 13.1 & 17.9 & 14.4 & 0.1 & 9.8 & 0.2 & 1.1 & 8.8 & 0.9 \\
        Random~\cite{straka2018sumeczech} & 11.7 & 15.5 & 12.7 & 0.1 & 2.0 & 0.1 & 0.7 & 10.3 & 0.8 \\
        Textrank~\cite{straka2018sumeczech} & 11.1 & 20.8 & 13.8 & 0.1 & 6.0 & 0.3 & 0.7 & 13.4 & 0.8 \\
        Tensor2Tensor~\cite{straka2018sumeczech} & 13.2 & 10.5 & 11.3 & 0.1 & 2.0 & 0.1 & 0.2 & 8.1 & 0.8 \\
        \bottomrule
    \end{tabular}
\end{table*}

\begin{table*}[ht!]
    \centering
    \caption{Results of implemented methods on the {\it POC-P} subset from Posel od \v{C}erchova dataset with precision (P), recall (R), and F1-score (F).}
    \label{tab:eval:poc-p}
    \begin{tabular}{
        l
        r
        r
        r
        r
        r
        r
        r
        r
        r
    }
        \toprule
        Method & \multicolumn{3}{c}{ROUGE\textsubscript{raw}-1} & \multicolumn{3}{c}{ROUGE\textsubscript{raw}-2} & \multicolumn{3}{c}{ROUGE\textsubscript{raw}-L} \\
        \cmidrule(lr){2-4} \cmidrule(lr){5-7} \cmidrule(lr){8-10}
        & {P} & {R} & {F} & {P} & {R} & {F} & {P} & {R} & {F} \\
        \midrule
        M7B-POC & \textbf{23.5} & {\bf 17.4} & {\bf 19.6} & \textbf{4.8} & {\bf 3.5} & \textbf{4.0} & \textbf{16.6} & {\bf 12.2} & \textbf{13.8} \\
        mT5-SC & 20.2 & 8.2 & 11.1 & 1.4 & 0.5 & 0.7 & 14.9 & 6.1 & 8.2 \\
        \bottomrule
    \end{tabular}
\end{table*}

\begin{table*}[ht!]
    \centering
    \caption{Results of implemented methods on {\it POC-I} subset from Posel od \v{C}erchova dataset with precision (P), recall (R), and F1-score (F).}
    \label{tab:eval:poc-i}
    \begin{tabular}{
        l
        r
        r
        r
        r
        r
        r
        r
        r
        r
    }
        \toprule
        Method & \multicolumn{3}{c}{ROUGE\textsubscript{raw}-1} & \multicolumn{3}{c}{ROUGE\textsubscript{raw}-2} & \multicolumn{3}{c}{ROUGE\textsubscript{raw}-L} \\
        \cmidrule(lr){2-4} \cmidrule(lr){5-7} \cmidrule(lr){8-10}
        & {P} & {R} & {F} & {P} & {R} & {F} & {P} & {R} & {F} \\
        \midrule
        M7B-POC & \textbf{19.3} & {\bf 17.6} & \textbf{18.0} & \textbf{3.2} & {\bf 2.8} & \textbf{2.9} & { 13.7} & {\bf 12.4} & \textbf{12.8} \\
        mT5-SC & 18.2 & 5.9 & 8.6 & 1.0 & 0.3 & 0.4 & \textbf{14.0} & 4.5 & 6.5 \\
        \bottomrule
    \end{tabular}
\end{table*}

\subsection{Results on the SumeCzech Dataset}

This experiment compares the results of the proposed mT5-SC and M7B-SC models with related work on the SumeCzech dataset, see Table~\ref{tab:eval:sc_scores_comparison}. 

The first comparative method, HT2A-S~\cite{Krotil2022TextSummarization}, is based on the mBART model, which is further fine-tuned on the SumeCzech dataset.
The other methods provided by the authors of the SumeCzech dataset~\cite{straka2018sumeczech} are as follows:  First, Random, Textrank and Tensor2Tensor~\cite{tensor2tensor}.

Table~\ref{tab:eval:sc_scores_comparison} demonstrates that the proposed M7B-SC method is very efficient, outperforming all other baselines and achieving new state-of-the-art results on this dataset.
Furthermore, the second proposed approach, mT5-SC, also performs remarkably well, consistently obtaining the second-best results.

\subsubsection{Results on Posel od \v{C}erchova Dataset}
This section evaluates the proposed methods on the Posel od \v{C}erchova dataset. Table~\ref{tab:eval:poc-p} shows the results on the {\it POC-P} subset containing summaries for every page (106 pages), while Table~\ref{tab:eval:poc-i} depicts the results on the {\it POC-I} subset, which is composed of the summaries of every article (25 issues).

These tables show clearly that, as in the previous case, M7T-POC model gives significantly better results than the mT5-SC model, and it is with a very high margin.

\section{\texorpdfstring{\uppercase{Conclusions}}{Conclusions}}
\label{sec:conclusion}
This paper explored the application of state-of-the-art large language models, specifically Mistral 7B and mT5, for summarization of Czech texts, addressing both modern and historical contexts. Our experiments demonstrated that the proposed M7B-SC model establishes a new benchmark for the SumeCzech dataset, achieving state-of-the-art performance, while the mT5-SC model also performed strongly, consistently ranking second.

Furthermore, we introduced a novel dataset, Posel od \v{C}erchova, dedicated for the summarization of historical Czech documents. By leveraging this dataset, we provided baseline results and highlighted the unique challenges posed by historical Czech texts.

These contributions not only advance the field of Czech text summarization but also pave the way for future research in processing historical documents, offering significant opportunities in cultural preservation and digital humanities. Future work could focus on further enhancing summarization quality, exploring hybrid modeling approaches, and extending the dataset for multilingual and cross-temporal studies.

\section*{\uppercase{Acknowledgements}}
This work was created with the partial support of the project R\&D of Technologies for Advanced Digitalization in the Pilsen Metropolitan Area (DigiTech) No. CZ.02.01.01/00/23\_021/0008436 and by the Grant No. SGS-2022-016 Advanced methods of data processing and analysis.
Computational resources were provided by the e-INFRA CZ project (ID:90254), supported by the Ministry of Education, Youth and Sports of the Czech Republic.

\bibliographystyle{apalike}
{\small
\bibliography{references}

\begin{thebibliography}{}

\bibitem[Ainslie et~al., 2023]{ainslie2023gqa}
Ainslie, J., Lee-Thorp, J., de~Jong, M., Zemlyanskiy, Y., Lebrón, F., and Sanghai, S. (2023).
\newblock Gqa: Training generalized multi-query transformer models from multi-head checkpoints.

\bibitem[Anthropic, 2024]{anthropic-2024}
Anthropic (2024).
\newblock {The Claude 3 Model Family: Opus, Sonnet, Haiku}.

\bibitem[Beltagy et~al., 2020]{beltagy2020longformer}
Beltagy, I., Peters, M.~E., and Cohan, A. (2020).
\newblock Longformer: The long-document transformer.

\bibitem[Christian et~al., 2016]{ChristianTFIDFSumm2016}
Christian, H., Agus, M., and Suhartono, D. (2016).
\newblock Single document automatic text summarization using term frequency-inverse document frequency (tf-idf).
\newblock {\em ComTech: Computer, Mathematics and Engineering Applications}, 7:285.

\bibitem[Cohan et~al., 2018]{cohan-etal-2018-discourse}
Cohan, A., Dernoncourt, F., Kim, D.~S., Bui, T., Kim, S., Chang, W., and Goharian, N. (2018).
\newblock A discourse-aware attention model for abstractive summarization of long documents.
\newblock In Walker, M., Ji, H., and Stent, A., editors, {\em Proceedings of the 2018 Conference of the North {A}merican Chapter of the Association for Computational Linguistics: Human Language Technologies, Volume 2 (Short Papers)}, pages 615--621, New Orleans, Louisiana. Association for Computational Linguistics.

\bibitem[Dettmers et~al., 2024]{qlora}
Dettmers, T., Pagnoni, A., Holtzman, A., and Zettlemoyer, L. (2024).
\newblock Qlora: efficient finetuning of quantized llms.
\newblock In {\em Proceedings of the 37th International Conference on Neural Information Processing Systems}, NIPS '23, Red Hook, NY, USA. Curran Associates Inc.

\bibitem[Devlin et~al., 2019]{devlin2019bert}
Devlin, J., Chang, M.-W., Lee, K., and Toutanova, K. (2019).
\newblock Bert: Pre-training of deep bidirectional transformers for language understanding.

\bibitem[Elman, 1990]{elman1990finding}
Elman, J.~L. (1990).
\newblock Finding structure in time.
\newblock {\em Cognitive Science}, 14(2):179--211.

\bibitem[Gholami et~al., 2021]{gholami2021survey}
Gholami, A., Kim, S., Dong, Z., Yao, Z., Mahoney, M.~W., and Keutzer, K. (2021).
\newblock A survey of quantization methods for efficient neural network inference.

\bibitem[Hasan et~al., 2021]{xlsumDataset}
Hasan, T. et~al. (2021).
\newblock Xlsum: A multilingual dataset for summarization.
\newblock In {\em Findings of the Association for Computational Linguistics: EMNLP 2021}, pages 2133--2149.

\bibitem[Hermann et~al., 2015]{hermann2015cnndm}
Hermann, K.~M., Kočiský, T., Grefenstette, E., Espeholt, L., Kay, W., Suleyman, M., and Blunsom, P. (2015).
\newblock Teaching machines to read and comprehend.
\newblock In {\em Advances in Neural Information Processing Systems (NeurIPS)}, pages 1693--1701.

\bibitem[Hu et~al., 2021]{hu2021lora}
Hu, E.~J., Shen, Y., Wallis, P., Allen-Zhu, Z., Li, Y., Wang, S., Wang, L., and Chen, W. (2021).
\newblock Lora: Low-rank adaptation of large language models.

\bibitem[Jiang et~al., 2023]{jiang2023mistral}
Jiang, A.~Q., Sablayrolles, A., Mensch, A., Bamford, C., Chaplot, D.~S., de~las Casas, D., Bressand, F., Lengyel, G., Lample, G., Saulnier, L., Lavaud, L.~R., Lachaux, M.-A., Stock, P., Scao, T.~L., Lavril, T., Wang, T., Lacroix, T., and Sayed, W.~E. (2023).
\newblock Mistral 7b.

\bibitem[Krotil, 2022]{Krotil2022TextSummarization}
Krotil, M. (2022).
\newblock Text summarization methods in czech.
\newblock Bachelor's thesis, Czech Technical University in Prague, Faculty of Electrical Engineering, Department of Cybernetics.

\bibitem[Kryscinski et~al., 2022]{kryscinski-etal-2022-booksum}
Kryscinski, W., Rajani, N., Agarwal, D., Xiong, C., and Radev, D. (2022).
\newblock {BOOKSUM}: A collection of datasets for long-form narrative summarization.
\newblock In Goldberg, Y., Kozareva, Z., and Zhang, Y., editors, {\em Findings of the Association for Computational Linguistics: EMNLP 2022}, pages 6536--6558, Abu Dhabi, United Arab Emirates. Association for Computational Linguistics.

\bibitem[Kudo and Richardson, 2018]{kudo-richardson-2018-sentencepiece}
Kudo, T. and Richardson, J. (2018).
\newblock {S}entence{P}iece: A simple and language independent subword tokenizer and detokenizer for neural text processing.
\newblock In Blanco, E. and Lu, W., editors, {\em Proceedings of the 2018 Conference on Empirical Methods in Natural Language Processing: System Demonstrations}, pages 66--71, Brussels, Belgium. Association for Computational Linguistics.

\bibitem[Lewis et~al., 2019]{lewis2019bart}
Lewis, M., Liu, Y., Goyal, N., Ghazvininejad, M., Mohamed, A., Levy, O., Stoyanov, V., and Zettlemoyer, L. (2019).
\newblock Bart: Denoising sequence-to-sequence pre-training for natural language generation, translation, and comprehension.

\bibitem[Lin, 2004]{lin-2004-rouge}
Lin, C.-Y. (2004).
\newblock {ROUGE}: A package for automatic evaluation of summaries.
\newblock In {\em Text Summarization Branches Out}, pages 74--81, Barcelona, Spain. Association for Computational Linguistics.

\bibitem[Liu et~al., 2020]{liu-etal-2020-multilingual-denoising}
Liu, Y., Gu, J., Goyal, N., Li, X., Edunov, S., Ghazvininejad, M., Lewis, M., and Zettlemoyer, L. (2020).
\newblock Multilingual denoising pre-training for neural machine translation.
\newblock {\em Transactions of the Association for Computational Linguistics}, 8:726--742.

\bibitem[Lloyd, 1982]{kmeans}
Lloyd, S. (1982).
\newblock Least squares quantization in pcm.
\newblock {\em IEEE Transactions on Information Theory}, 28(2):129--137.

\bibitem[Loshchilov and Hutter, 2017]{loshchilov2017decoupled}
Loshchilov, I. and Hutter, F. (2017).
\newblock Decoupled weight decay regularization.
\newblock {\em arXiv preprint arXiv:1711.05101}.

\bibitem[Mihalcea and Tarau, 2004]{mihalcea-tarau-2004-textrank}
Mihalcea, R. and Tarau, P. (2004).
\newblock {T}ext{R}ank: Bringing order into text.
\newblock In Lin, D. and Wu, D., editors, {\em Proceedings of the 2004 Conference on Empirical Methods in Natural Language Processing}, pages 404--411, Barcelona, Spain. Association for Computational Linguistics.

\bibitem[Miller, 2019]{miller2019leveraging}
Miller, D. (2019).
\newblock Leveraging bert for extractive text summarization on lectures.

\bibitem[Nallapati et~al., 2017]{nallapati2017summarunner}
Nallapati, R., Zhai, F., and Zhou, B. (2017).
\newblock Summarunner: A recurrent neural network based sequence model for extractive summarization of documents.
\newblock In {\em Proceedings of the Thirty-First AAAI Conference on Artificial Intelligence (AAAI)}, pages 3075--3081.

\bibitem[Narayan et~al., 2018]{xsumDataset}
Narayan, S., Cohen, S.~B., and Lapata, M. (2018).
\newblock Extreme summarization (xsum).
\newblock In {\em Proceedings of the 2018 Conference on Empirical Methods in Natural Language Processing}, pages 931--936.

\bibitem[OpenAI, 2024]{openai2024gpt4}
OpenAI (2024).
\newblock Gpt-4 technical report.

\bibitem[Page et~al., 1999]{Page1999ThePC}
Page, L., Brin, S., Motwani, R., and Winograd, T. (1999).
\newblock The pagerank citation ranking : Bringing order to the web.
\newblock In {\em The Web Conference}.

\bibitem[Raffel et~al., 2020a]{2020t5}
Raffel, C., Shazeer, N., Roberts, A., Lee, K., Narang, S., Matena, M., Zhou, Y., Li, W., and Liu, P.~J. (2020a).
\newblock Exploring the limits of transfer learning with a unified text-to-text transformer.
\newblock {\em Journal of Machine Learning Research}, 21(140):1--67.

\bibitem[Raffel et~al., 2020b]{raffel2020t5}
Raffel, C., Shazeer, N., Roberts, A., Lee, K., Narang, S., Matena, M., Zhou, Y., Li, W., and Liu, P.~J. (2020b).
\newblock Exploring the limits of transfer learning with a unified text-to-text transformer.
\newblock {\em Journal of Machine Learning Research}, 21(140):1--67.

\bibitem[Scialom et~al., 2020]{mlsumDataset}
Scialom, T. et~al. (2020).
\newblock Mlsum: Multilingual summarization dataset.
\newblock In {\em Proceedings of the 2020 Conference on Empirical Methods in Natural Language Processing}, pages 2146--2161.

\bibitem[Straka et~al., 2018]{straka2018sumeczech}
Straka, M., Mediankin, N., Kocmi, T., {\v{Z}}abokrtsk{\'y}, Z., Hude{\v{c}}ek, V., and Haji{\v{c}}, J. (2018).
\newblock {S}ume{C}zech: Large {C}zech news-based summarization dataset.
\newblock In {\em Proceedings of the Eleventh International Conference on Language Resources and Evaluation ({LREC} 2018)}, Miyazaki, Japan. European Language Resources Association (ELRA).

\bibitem[Straka and Strakov{\'a}, 2018]{straka2018rougeraw}
Straka, M. and Strakov{\'a}, J. (2018).
\newblock Rougeraw: Language-agnostic evaluation for summarization.
\newblock {\em Proceedings of the International Conference on Computational Linguistics}.

\bibitem[Vaswani et~al., 2018]{tensor2tensor}
Vaswani, A., Bengio, S., Brevdo, E., Chollet, F., Gomez, A.~N., Gouws, S., Jones, L., Kaiser, L., Kalchbrenner, N., Parmar, N., Sepassi, R., Shazeer, N., and Uszkoreit, J. (2018).
\newblock Tensor2tensor for neural machine translation.
\newblock {\em CoRR}, abs/1803.07416.

\bibitem[Vaswani et~al., 2017]{vaswani2017attention}
Vaswani, A., Shazeer, N., Parmar, N., Uszkoreit, J., Jones, L., Gomez, A.~N., Kaiser, L.~u., and Polosukhin, I. (2017).
\newblock Attention is all you need.
\newblock In Guyon, I., Luxburg, U.~V., Bengio, S., Wallach, H., Fergus, R., Vishwanathan, S., and Garnett, R., editors, {\em Advances in Neural Information Processing Systems}, volume~30. Curran Associates, Inc.

\bibitem[Wolf et~al., 2020]{wolf-etal-2020-transformers}
Wolf, T., Debut, L., Sanh, V., Chaumond, J., Delangue, C., Moi, A., Cistac, P., Rault, T., Louf, R., Funtowicz, M., Davison, J., Shleifer, S., von Platen, P., Ma, C., Jernite, Y., Plu, J., Xu, C., Scao, T.~L., Gugger, S., Drame, M., Lhoest, Q., and Rush, A.~M. (2020).
\newblock Transformers: State-of-the-art natural language processing.
\newblock In {\em Proceedings of the 2020 Conference on Empirical Methods in Natural Language Processing: System Demonstrations}, pages 38--45, Online. Association for Computational Linguistics.

\bibitem[Xue et~al., 2021a]{xue-etal-2021-mc4}
Xue, L., Constant, N., Roberts, A., Kale, M., Al-Rfou, R., Siddhant, A., Barua, A., and Raffel, C. (2021a).
\newblock {mC4}: A massively multilingual cleaned crawl corpus.
\newblock In {\em Proceedings of the 2021 Conference on Empirical Methods in Natural Language Processing (EMNLP)}, pages 7517--7532, Online and Punta Cana, Dominican Republic. Association for Computational Linguistics.

\bibitem[Xue et~al., 2021b]{xue-etal-2021-mt5}
Xue, L., Constant, N., Roberts, A., Kale, M., Al-Rfou, R., Siddhant, A., Barua, A., and Raffel, C. (2021b).
\newblock m{T}5: A massively multilingual pre-trained text-to-text transformer.
\newblock In Toutanova, K., Rumshisky, A., Zettlemoyer, L., Hakkani-Tur, D., Beltagy, I., Bethard, S., Cotterell, R., Chakraborty, T., and Zhou, Y., editors, {\em Proceedings of the 2021 Conference of the North American Chapter of the Association for Computational Linguistics: Human Language Technologies}, pages 483--498, Online. Association for Computational Linguistics.

\bibitem[Yang et~al., 2016]{yang2016hierarchical}
Yang, Z., Yang, D., Dyer, C., He, X., Smola, A., and Hovy, E. (2016).
\newblock Hierarchical attention networks for document classification.
\newblock In {\em Proceedings of the 2016 Conference of the North American Chapter of the Association for Computational Linguistics: Human Language Technologies}, pages 1480--1489.

\bibitem[Zhang et~al., 2019]{zhang2019pegasus}
Zhang, J., Zhao, Y., Saleh, M., and Liu, P.~J. (2019).
\newblock Pegasus: Pre-training with extracted gap-sentences for abstractive summarization.

\end{thebibliography}
}

\end{document}